\documentclass[letterpaper]{article} 
\usepackage{aaai2026}  
\usepackage{times}  
\usepackage{helvet}  
\usepackage{amsmath} 
\usepackage{mathtools} 
\usepackage{booktabs}
\usepackage{multirow}
\usepackage{array} 
\usepackage{courier}  
\usepackage[hyphens]{url}  
\usepackage{graphicx} 
\usepackage{natbib}  
\usepackage{caption} 
\usepackage{algorithm}
\usepackage{algorithmic}
\usepackage{amssymb} 
\usepackage{newfloat}
\usepackage{listings}
\newcommand{\etal}{\textit{et al.}}
\DeclareCaptionStyle{ruled}{labelfont=normalfont,labelsep=colon,strut=off} 
\floatstyle{ruled}
\newfloat{listing}{tb}{lst}{}
\floatname{listing}{Listing}
\title{AMMKD: Adaptive Multimodal Multi-teacher Distillation \\ for Lightweight Vision-Language Models}
\author{
Yuqi Li\textsuperscript{1} 
Chuanguang Yang\textsuperscript{1}\thanks{Corresponding Author.}
Junhao Dong\textsuperscript{2},
Zhengtao Yao\textsuperscript{3}, 
Haoyan Xu\textsuperscript{3} 
Zeyu Dong\textsuperscript{1}, \\
Hansheng Zeng\textsuperscript{4}, 
Zhulin An\textsuperscript{1}\footnotemark[1], 
Yingli Tian\textsuperscript{5}
}

\affiliations{
    \textsuperscript{1}Institute of Computing Technology, Chinese Academy of Sciences \\
    \textsuperscript{2}Nanyang Technological University,
    \textsuperscript{3}University of Southern California, \\
    \textsuperscript{4}University of Hong Kong,
    \textsuperscript{5}The City University of New York
}

\usepackage{bibentry}

\begin{document}

\maketitle

\begin{abstract}
The success of large-scale visual language pretraining (VLP) models has driven widespread adoption of image-text retrieval tasks. However, their deployment on mobile devices remains limited due to large model sizes and computational complexity. We propose \textbf{A}daptive \textbf{M}ulti-Modal \textbf{M}ulti-Teacher \textbf{K}nowledge \textbf{D}istillation (AMMKD), a novel framework that integrates multi-modal feature fusion, multi-teacher distillation, and adaptive optimization to deliver lightweight yet effective retrieval models. Specifically, our method begins with a feature fusion network that extracts and merges discriminative features from both the image and text modalities. To reduce model parameters and further improve performance, we design a multi-teacher knowledge distillation framework to pre-train two CLIP teacher models. We decouple modalities by pre-computing and storing text features as class vectors via the teacher text encoder to enhance efficiency. To better align teacher and student outputs, we apply KL scatter for probability distribution matching. Finally, we design an adaptive dynamic weighting scheme that treats multi-teacher distillation as a multi-objective optimization problem. By leveraging gradient space diversity, we dynamically adjust the influence of each teacher, reducing conflicts and guiding the student toward more optimal learning directions. Extensive experiments on three benchmark datasets demonstrate that AMMKD achieves superior performance while significantly reducing model complexity, validating its effectiveness and flexibility.
\end{abstract}

\section{Introduction}

Multimodal modeling focuses on learning how to represent and summarize multimodal data in a way that exploits the complementarity and redundancy of multiple modalities. The heterogeneity of multimodal data makes it challenging to construct such representations. For example, language is often symbolic, while audio and visual forms will be represented as signals. Unimodal representations are responsible for representing information as numerical vectors that computers can process or further abstract into higher-level feature vectors, whereas multimodal representations refer to learning better feature representations by taking advantage of complementarities between multiple modalities and eliminating redundancies between modalities.
\begin{figure}[!t]
	\centering
	
	\includegraphics[width=\linewidth]{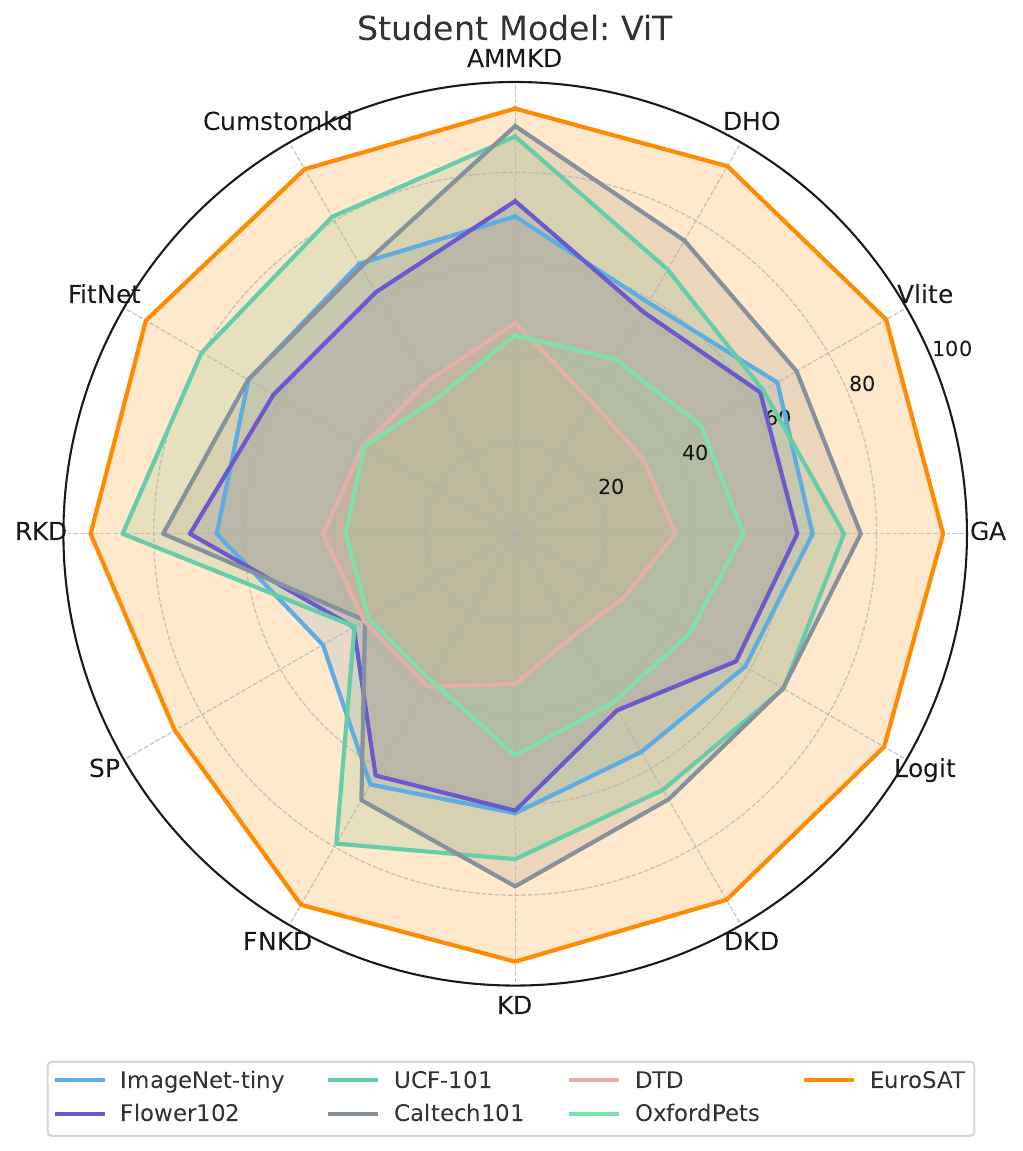}
	
	\caption{Migration performance of 12 knowledge distillation methods (Cumstomkd\cite{lee2025customkd}, FitNet\cite{romero2014fitnets}, RKD\cite{park2019relational}, SP\cite{tung2019similarity}, FNKD\cite{xu2020feature}, KD\cite{hinton2015distilling}, DKD\cite{zhao2022decoupled}, Logit\cite{sun2024logit}, GA\cite{wang2018adversarial}, Vlite\cite{jang2025vl2lite}, DHO\cite{kang2025simple},AMMKD) on the two backbone networks of ResNet and Vision Transformer (ViT) Covers seven different data sets (ImageNet - tiny/Flower102 / UCF - 101 / Caltech101 / DTD/OxfordPets/EuroSAT).}
	\label{fig:radar}
\end{figure}

CLIP~\cite{radford2021learning} has taken a significant step forward in the multimodal domain, with the main contribution being the use of unsupervised textual information as a supervised signal to learn visual features. This training is performed via the contrast loss method to achieve uniform embedding space learning of multimodal signals, thus efficiently aligning the representations of different modalities. This approach not only improves the performance of the model in multimodal tasks but also provides important directions for future research. Due to the significant gaps between different modalities, multimodal data not only provides effective information, but also brings higher information redundancy. However, most existing fusion methods focus mainly on simply integrating multimodal features or employing a single optimization strategy during joint training, which usually leads to unsatisfactory performance.

Knowledge distillation (KD)~\cite{xu2024survey} can be categorized into single-teacher and multi-teacher paradigms. The single-teacher setting transfers knowledge from a single model to a student, but the teacher’s capacity inherently limits the student’s performance. In contrast, multi-teacher distillation aggregates knowledge from multiple teachers, providing richer supervision and improved generalization.

However, vanilla multi-teacher distillation often treats each teacher equally or fuses their outputs naively, without considering their expertise, modality relevance, or potential conflicts. This may lead to ineffective or even contradictory guidance for the student model, limiting its performance. Recent studies~\cite{zhang2023adaptive} propose adaptive strategies to address these challenges by weighting or selecting teacher knowledge more intelligently.


\begin{figure*}[t]
	\centering
	\includegraphics[width=\linewidth]{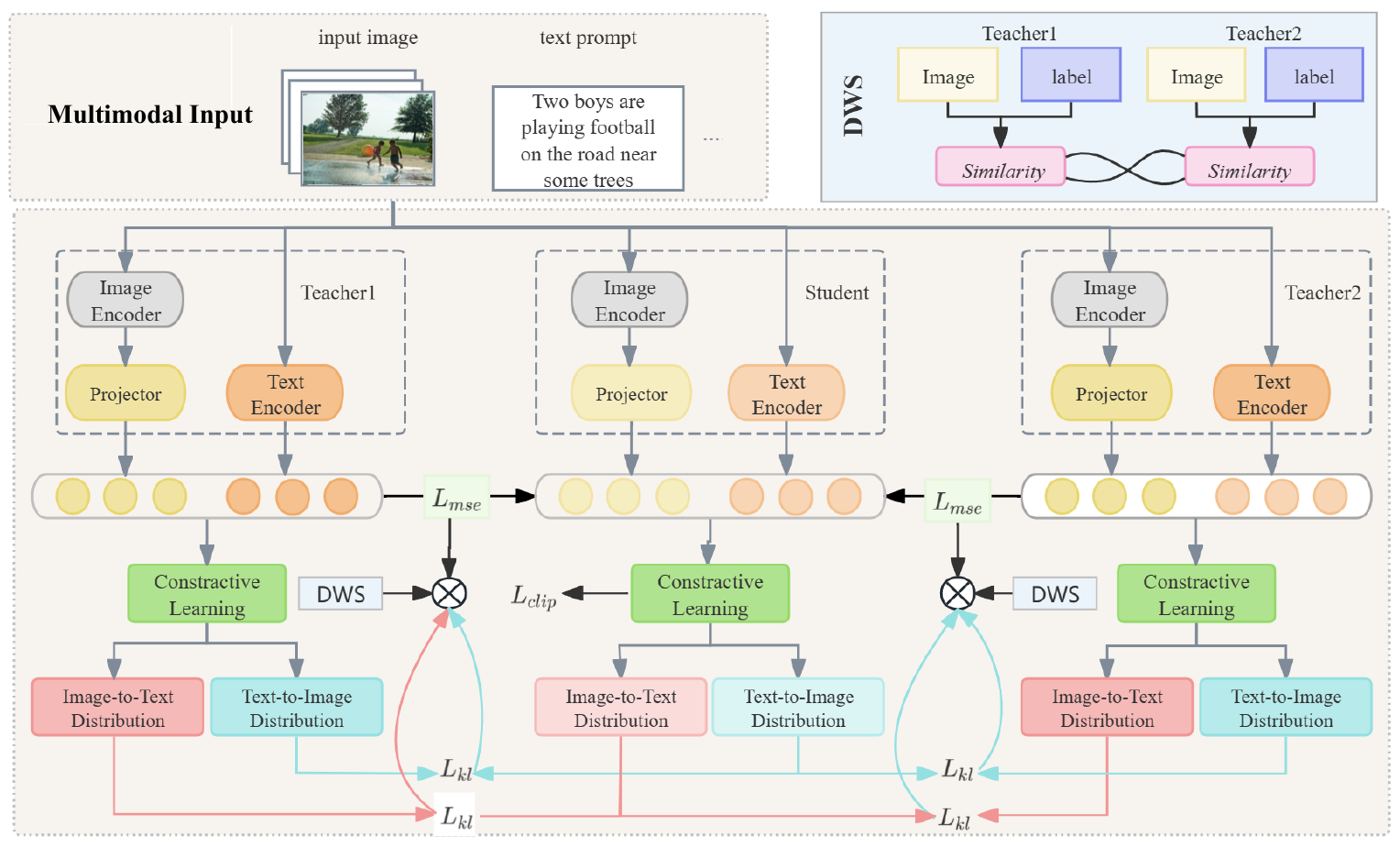}
	\caption{Framework of the Adaptive Multi-Teacher Multi-Modal Knowledge Distillation (AMMKD) method. It contains one multimodal student model and two pre-trained multimodal teacher models. The student and the teachers process pairs of images and text simultaneously. During the student's training step, the parameters of the pre-trained teachers are frozen and the multimodal knowledge extracted from the teachers is aggregated to guide the student's training.}
	\label{fig:framework}
\end{figure*}


In the practical operation of the distillation process, knowledge transfer is usually achieved through a simple method of averaging the knowledge or information extracted from each collection of teacher models. However, this approach fails to adequately take into account differences in diversity across teacher models within the set. As a result, competing and interfering noises may be generated during the distillation process, thus affecting the overall effect.

Facing these challenges, we propose an adaptive multimodal multi-teacher knowledge distillation framework, as shown in Figure \ref{fig:framework}. This can better alleviate the problem of large parameter scales of multimodal models. It can keep the student model as close as possible to the teacher model in terms of performance, while drastically reducing the number of parameters of the model, to satisfy the needs in graphic retrieval, graphic QA, pure text comprehension, and pure visual comprehension tasks. To further improve the accuracy of the distilled student model, it is crucial to take full advantage of the integration of the multi-teacher model while minimizing the possible adverse effects among the teacher models. Therefore, in the framework of multi-teacher distillation, each teacher model should be provided with a weight to help the student model learn and absorb knowledge from different teachers more comprehensively. 
	
	

To summarize, our main contributions are as follows:
\\1. We propose a multimodal multi-teacher knowledge distillation framework that leverages guidance from multiple teachers to improve the cross-modal retrieval performance of a lightweight CLIP model.
\\2. We introduce a novel loss function designed to align the text and image modalities of the teacher model with those of the student model, respectively, thereby enhancing modality-specific knowledge transfer.
\\3. To address the potential presence of noisy or weak-performing teachers, we present a multi-teacher dynamic weighting strategy. This strategy employs a dual optimization algorithm to weigh and extract useful knowledge from all available teachers adaptively.
\\4. Extensive experiments on several public multimodal datasets demonstrate that our AMMKD significantly outperforms traditional single-teacher distillation baseline approaches.

\section{Related Work}

\subsection{Multi-teacher Knowledge Distillation}

Within the framework of multi-teacher knowledge distillation, each teacher network offers soft-label predictions, which are then combined to direct the student network's learning. AVER \cite{wang2022efficient} adopts a straightforward approach by assigning equal weight to each teacher's prediction, treating all contributions as equally valuable. In contrast, RLKD \cite{yuan2021reinforced} introduces reinforcement learning to improve the selection process, filtering out less suitable teachers before averaging the remaining predictions. This refinement aims to reduce the influence of teachers whose guidance may be sub-optimal for the student. Other methods focus on tailoring the weight each teacher contributes based on their unique qualities. For example, EBKD \cite{kwon2020adaptive} employs information entropy to assign distinct weights to each teacher, adapting based on the diversity of predictions. Similarly, CA-MKD \cite{zhang2022confidence} evaluates the reliability of individual teacher predictions by calculating cross-entropy with ground-truth labels, ensuring that teachers with higher predictive confidence have a greater impact on the student's learning. On the other hand, AEKD \cite{du2020agree} takes an entirely different perspective by examining teacher diversity within the gradient space, which helps to understand how variations among teachers contribute to the student's training process. Some multi-teacher KD techniques maintain fixed or uniform weights for simplicity \cite{you2017learning, fukuda2017efficient, wu2019multi}, while Yuang \etal \cite{liu2020adaptive} propose to leverage latent factors to calculate dynamic weights, adjusting each teacher's influence according to underlying factors in the data. Despite their benefits, these label-independent weighting strategies present some limitations. For example, entropy-based weighting \cite{kwon2020adaptive} can inadvertently prioritize predictions with low variance, which may not always be reliable. Optimization-based strategies, on the other hand, may rely too heavily on majority consensus, making them susceptible to noisy data that could mislead the student \cite{du2020agree}. With a meta-weight network, Adaptive Multi-Teacher Knowledge Distillation with Meta-Learning (MMKD) \cite{zhang2023adaptive} proposes to adaptively integrate knowledge from multiple teachers and enhance student model performance by considering both teacher diversity and knowledge compatibility.

\subsection{Multi-modal Knowledge Distillation}

Multi-Modal Knowledge Distillation (MKD) has emerged as a powerful technique aimed at improving knowledge transfer across various modalities, including images \cite{li2023curriculum, li2021online, yang2022mixskd}, text \cite{jiao2019tinybert, chen2021simplified, zheng2024microbert}, and audio \cite{schmid2023efficient, choi2022temporal}. By integrating information from these diverse sources, MKD leverages the complementary strengths of each modality, thereby enhancing the performance of student models in tasks that necessitate a multi-faceted understanding and representation. Wang \etal \cite{wang2022multimodal} enhance the generalization capabilities of models on multi-modal tasks by adapting distillation strategies that exploit the complementarity of different modalities, which improves the performance of individual encoders through cross-modal knowledge transfer. Additionally, Chen \etal \cite{liang2024module} introduce Module-wise Adaptive Distillation, explicitly designed for multimodal foundation models, allowing for knowledge distillation on a module-level basis. Aiming at achieving an alignment of the feature maps and attention distributions between the teacher and student models, DistillVLM \cite{fang2021compressing} yields a more compact yet effective visual-linguistic compression technique. CLIP \cite{radford2021learning} demonstrates that models can effectively learn to associate images with text descriptions without the need for explicit pairings. By utilizing natural language as a supervisory signal, CLIP produces a robust zero-shot image classification model that generalizes across a wide array of visual concepts. Building on this foundation, TinyCLIP \cite{wu2023tinyclip} aims to mimic the affinity between teacher and student models while inheriting weights to develop a compact yet practical student model. Furthermore, CLIP-KD \cite{yang2024clipkd} explores strategies for distilling large CLIP models into smaller, more efficient student models, focusing on techniques such as feature mimicry and interactive contrastive learning to bolster the student model's generalization capabilities and performance on zero-shot tasks.
\section{Methodology}
\subsection{Preliminary}
The general framework of the method in this paper is shown in Figure \ref{fig:framework} and contains two pre-trained multimodal teacher models and one multimodal student model. We input $M$ paired image and text datasets $X=\{(P_{i},Q_{i})\}_{i=1}^{M}$, where $p_i\in\mathbb{R}^{C\times H\times W}$ denotes the $i$th image sample with channel $C$, width $W$ and height $H$, and ${{Q}_{i}}$ is the $i$th text sample. The model in our framework utilizes an image feature extraction module and a text feature extraction module to extract rich information from the image and text samples, respectively. Then, image features and text features are mapped into a vector space for comparative learning using a projector to obtain the Image-to-Text probability distribution and the Text-to-Image probability distribution, respectively. This network is trained in two steps. First, the teacher network is trained using the cross-entropy loss function to obtain the CLIP teacher model. Then, the parameters of the pre-trained teachers are frozen, and the multimodal knowledge imparted by the two teachers is used to guide the training of the multimodal students. The key elements are described in detail in the following subsections.

\subsection{Multi-Modal Feature Extraction}
In this paper, CLIP is used as a base model.CLIP consists of two encoders, one for images and the other for text. Given a labeled visual recognition dataset $X=\{(P_{i},Q_{i})\}_{i=1}^{M}$, which includes a set of $N$ class names $c={{\{{{c}_{j}}\}}_{j=1}^{N}}$, CLIP uses a template to generate a description ${{Q}_{i}}$ as “a photo of a $\left \{{{{c}_{j}}}\right \}$”.Each text description ${{Q}_{i}}$ is then input to a text encoder ${{f}_{T}}$ to obtain normalized text features ${{w}_{i}}=\frac{{{f}_{T}}({{Q}_{i}})}{||{{f}_{T}}({{Q}_{i}})|{{|}_{2}}}\in{{R}^{d}}$ , where $d$ denotes the feature dimension. Similarly, an image ${{P}_{i}}$ is input to the image encoder ${{f}_{I}}$ which takes it as input and generates normalized image features ${{u}_{i}}=\frac{{{f}_{I}}({{P}_{i}})}{{{||{{f}_{I}}({{P}_{i}})||}_{2}}}\in{{R}^{d}}$.The image-to-text output probability is calculated as follows:

\begin{equation}
p(P|Q)=\frac{exp({{u}_{i}}\cdot{{w}_{i}}/\tau)}{\sum_{j=1}^{N}{exp({{u}_{i}}\cdot{{w}_{j}}/\tau)}}.
\end{equation}
CLIP conducts a symmetric image-text alignment. Given the text embedding  as the anchor, the text-to-image output probability is formulated as:
\begin{equation}
p(Q|P)=\frac{exp({{w}_{i}}\cdot{{u}_{i}}/\tau)}{\sum_{j=1}^{N}{exp({{w}_{i}}\cdot{{u}_{j}}/\tau)}},
\end{equation}
where $\tau$ is the temperature parameter.
\subsection{Multi-Teacher Multi-Modal Knowledge Distillation}

After identifying the multimodal student network, we propose a novel multimodal multi-teacher knowledge distillation to comprehensively aggregate the knowledge of multiple pre-trained multimodal teachers, thereby effectively training the multimodal student to achieve better performance. The overall loss function of the student consists of distillation losses (i.e., ${{L}_{kl}}$ and ${{L}_{mse}}$) and student loss (i.e., ${{L}_{clip}}$). The teacher and student image-to-text contrastive distributions ${{p}_{k}^{T}}(P|Q)$and ${{p}^{S}}(P|Q)$
 are formulated as:
\begin{equation}
{{p}_{k}^{T}}(P|Q)=\frac{exp({{u}_{i}^{T}}\cdot{{w}_{i}^{T}}/{{\tau}^{T}})}{\sum_{j=1}^{N}{exp({{u}_{i}^{T}}\cdot{{w}_{j}^{T}}/{{\tau}^{T}})}},
\end{equation}

\begin{equation}
{{p}^{S}}(P|Q)=\frac{exp({{u}_{i}^{S}}\cdot{{w}_{i}^{S}}/{{\tau}^{S}})}{\sum_{j=1}^{N}{exp({{u}_{i}^{S}}\cdot{{w}_{j}^{S}}/{{\tau}^{S}})}}.
\end{equation}
Similarly, text-to-image contrastive distributions  ${{p}_{k}^{T}}(P|Q)$and ${{p}^{S}}(P|Q)$ as follows:
\begin{equation}
{{p}^{S}}(P|Q)=\frac{exp({{u}_{i}^{S}}\cdot{{w}_{i}^{S}}/{{\tau}^{S}})}{\sum_{j=1}^{N}{exp({{u}_{i}^{S}}\cdot{{w}_{j}^{S}}/{{\tau}^{S}})}},
\end{equation}

\begin{equation}
{{p}^{S}}(P|Q)=\frac{exp({{u}_{i}^{S}}\cdot{{w}_{i}^{S}}/{{\tau}^{S}})}{\sum_{j=1}^{N}{exp({{u}_{i}^{S}}\cdot{{w}_{j}^{S}}/{{\tau}^{S}})}}.
\end{equation}
This paper aligns the contrast distribution between teacher and student by KL divergence loss. For image to text and text to image, the distillation loss for the $k$ teacher and the student loss ${{L}_{clip}}$ can be calculated as follows:

\begin{equation}
{{L}_{kl}}(P|Q)=KL({{p}_{k}^{T}}(P|Q),{{p}^{S}}(P|Q)),
\end{equation}

\begin{equation}
{{L}_{kl}}(Q|P)=KL({{p}_{k}^{T}}(Q|P),{{p}^{S}}(Q|P)),
\end{equation}

\begin{equation}
{{L}_{clip}}=CE\left ({y,\cos{\left ({{{u}_{i}},{{w}_{i}}}\right )}}\right ),
\end{equation}
where $KL$ (·) is the Kullback-Leibler divergence function [25]. $CE$ (·) represents the contrast learning loss function, and $y$ represents the label. $cos$(·) represents the cosine similarity calculation.

Align feature vectors between teachers and students to directly reduce knowledge gaps. Intuitively, if student characteristics could be perfectly aligned with teacher characteristics, their performance gaps would disappear. This work uses Mean Squared Error (MSE) to guide students to imitate the teacher's image and text feature vectors:
\begin{equation}
{{L}_{mse}}=MSE({{u}_{i}^{T}},{{u}_{i}^{S}})+MSE({{w}_{i}^{T}},{{w}_{i}^{S}}).
\end{equation}
By combining Equations (7), (8), (9), and (10), we can get the overall loss function of the student model as follows:
\begin{equation}
L={{L}_{kl}}+{{L}_{clip}}+{{L}_{mse}},
\end{equation}

\begin{equation}
{{L}_{kl}}={{\lambda}_{1}}{{L}_{kl}}(P|Q)+{{\lambda}_{2}}{{L}_{kl}}(Q|P),
\end{equation}
where ${\lambda}_{1}$ and ${\lambda}_{2}$ are the dynamic parameters that balance the influence of teacher loss, and the combination strategy of the teacher model can be flexibly optimized. In essence, this parameter-guided method introduces an intelligent regulation mechanism in the collaborative training of multi-teacher models, so that the weight allocation of teacher models can be continuously optimized with the deepening of the training process to meet the learning needs of student models. The details are explained in the next section.

\begin{table*}[!t]
\centering
\setlength{\tabcolsep}{3pt} 
\caption{Methods Performance Comparison}
\begin{tabular}{c *{7}{cc}} 
\toprule
\multirow{2}{*}{Methods} & 
\multicolumn{2}{c}{ImageNet-tiny} & 
\multicolumn{2}{c}{Flower102} & 
\multicolumn{2}{c}{UCF-101} & 
\multicolumn{2}{c}{Caltech101} & 
\multicolumn{2}{c}{DTD} & 
\multicolumn{2}{c}{OxfordPets} & 
\multicolumn{2}{c}{EuroSAT} \\
\cmidrule(lr){2-3} \cmidrule(lr){4-5} \cmidrule(lr){6-7} \cmidrule(lr){8-9} \cmidrule(lr){10-11} \cmidrule(lr){12-13} \cmidrule(l){14-15}
& Resnet & Vit & Resnet & Vit & Resnet & Vit & Resnet & Vit & Resnet & Vit & Resnet & Vit & Resnet & Vit \\
\midrule
Cumstomkd & 69.01 & 69.01 & 58.00 & 61.72 & 72.71 & 81.01 & 80.16    & 68.21    & 33.25    & 38.92    & 49    & 34.67    & 94.91    & 93.19    \\
FitNet  & 68.95 & 68.17 & 59.10 & 61.72 & 73.16 & 80.16 & 80.55    & 68.25    & 33.51    & 38.83    & 49.17    & 38.58    & 94.87    & 94.39    \\
RKD     & 69.08 & 66.06 & 62.21 & 71.98 & 80.04 & 86.89 & 81.67    & 77.92    & 31.12    & 42.55    & 47.42    & 37.50   & 91.46    & 93.98    \\
SP      & 52.46 & 49.15 & 32.48 & 41.27 & 72.86 & 41.06 & 74.89    & 38.37    & 19.77    & 38.56    & 47.83    & 37.50    & 75.98    & 87.02    \\
FNKD    & 66.46 & 64.09 & 55.68 & 61.78 & 69.94 & 79.19 & 76.08    & 68.08    & 29.88    & 39.01    & 48.08    & 37.42    & \textbf{95.43}    & \textbf{94.83}    \\
KD      & 60.25 & 61.85 & 58.55 & 61.29 & 72.64 & 72.02 & 79.99    & 78.04    & 32.80    & 33.28    & 48.67    & 49.06    & 94.56    & 94.70    \\
DKD     & 62.12 & 55.78 & 52.35 & 45.12 & 69.01 & 65.43 & 72.46    & 67.89    & 29.57    & 25.98    & 47.68    & 43.21    & 93.03    & 93.57    \\
Logit   & 60.29 & 58.79 & 49.88 & 56.47 & 67.23 & 68.60 & 70.62    & 68.46    & 27.31    & 27.95   & 47.50    & 44.47    & 95.26    & 94.30    \\
GA      & 68.19 & 65.85 & 58.12 & 62.45 & 71.60 & 72.80 & 79.99    & 76.49    & 32.54    & 35.49    & \textbf{49.92}    & \textbf{50.50}    & 94.67    & 94.72    \\
Vlite   & 68.76 & 67.00 & 56.84 & 62.64 & 65.94 & 63.42 & 72.03    & 71.94    & 27.57    & 32.79    & 47.00    & 47.64    & 95.39    & 94.82    \\
DHO     & 61.81 & 59.06 & 57.63 & 56.71 & 70.27 & 67.53 & 76.47    & 74.94    & 29.70    & 33.81    & 46.33    & 44.55    & 92.98    & 93.96    \\
\midrule
\textbf{AMMKD(OURS)}   & \bfseries69.53 & \bfseries70.25 & \bfseries74.56 & \bfseries73.63 & \bfseries86.45 & \bfseries87.93 & \bfseries91.71    & \bfseries90.26    & \bfseries46.22    & \bfseries46.63    & 46.71    & 43.92    & 94.57    & 94.11    \\
\bottomrule

\end{tabular}
\label{result}
\vspace{-0.2cm}
\end{table*}

\subsection{Multi-Teacher Dynamic Weight Selection}
As shown in Equation (12), if the losses are added with a parameter of 1, all teacher losses are considered equally. However, since teachers may offer different learning directions to students, conflicts and competition may arise among teachers. Therefore, the final direction of learning will be determined by the primary teacher, thus weakening the guidance of other teachers. To this end, we propose a multi-teacher dynamic selection mechanism, which can be regarded as a dynamic weighting strategy that can adaptively extract knowledge from all teachers.

First, the similarity between each teacher image and the current label text is taken as a dynamic weight, denoted as ${r}_{1}$ and ${r}_{2}$, as shown in the module of Figure \ref{fig:framework} DWS. The resulting formula is shown in Equation (12), where:
\begin{equation}
{{\lambda}_{1}}=\frac{{{r}_{1}}}{{{r}_{1}}+{{r}_{2}}},
\end{equation}

\begin{equation}
{{\lambda}_{2}}=\frac{{{r}_{2}}}{{{r}_{1}}+{{r}_{2}}}.
\end{equation}
Second, when faced with internal incoherence in multi-teacher knowledge distillation (KD), we choose to re-imagine multi-teacher knowledge distillation as a multi-objective optimization (MOO) framework from the perspective of gradient space. This change aims to consider the knowledge contribution of multiple teacher models simultaneously, in order to find a student model configuration that can harmoniously integrate the advantages of each teacher. Next, we will elaborate on how to use the multi-gradient descent algorithm (MGDA) to find the optimal solution in this complex multi-objective environment.

On the basis of MGDA, it is assumed that for each task, KKT (Kuarush-Huhn-Tucker, a necessary condition for optimal solution of nonlinear programming) holds:

There is ${{\alpha}_{1}},\dots,{{\alpha}_{K}}\geq0$, s.t. $\sum_{k=1}^{K}{{{\alpha}_{k}}}=1$,$\sum_{k=1}^{K}{{{\alpha}_{k}}}{{\nabla}_{p}}CE\left ({{{p}^{S}},{{p}_{k}^{T}}}\right )=0$. In the shared layer, there is no feasible direction of decline; that is, it belongs to the global best,  and the optimal solution for all teachers is found.

For any teacher $k$, ${{\nabla}_{p}}CE\left ({{{p}^{S}},{{p}_{k}^{T}}}\right )=0$.

Any solution satisfying the above conditions is a Pareto stationary point, and every optimization solution is a stationary point, and vice versa is not true, so we rewrite the optimization problem as follows:

\begin{equation}
\begin{aligned}
\min_\alpha \ & \frac{1}{2} \Bigg\| \sum_{k=1}^K \alpha_k \nabla_p \big( g(P|Q) + g(Q|P) \big) \Bigg\|^2, \\
\text{s.t.} \ & \sum_{k=1}^K \alpha_k = 1, \ \alpha_k \geq 0, \ \forall k \in [1{:}K],
\end{aligned}
\end{equation}
\noindent
where $g(P|Q)=CE(p^S(P|Q),p^T(P|Q))$and$g(Q|P)=CE(p^S(Q|P),p^T(Q|P))$
Represent the image-to-text cross entropy and the text-to-image cross entropy of the student and teacher models, respectively, and are gradient steps. In this way, the optimization direction will be less affected by those deviating or noisy teachers, thus reducing performance degradation. Our approach can be viewed as a dynamic weighting strategy for each teacher to refine knowledge during the adaptive training of the student network.

\begin{table*}[t]
    \caption{Performance on Flower102 and UCF-101, where ``MT'' denotes ``multi-teacher'' and ``DWS'' represents multi-teacher dynamic weight selection.}
    \centering
    \resizebox{\textwidth}{!}{ 
    \begin{tabular}{@{}l|cccccc|cc@{}}
        \toprule
        & \multicolumn{1}{c}{$L_{CLIP}$} 
        & \multicolumn{1}{c}{$L_{KL}$} 
        & \multicolumn{1}{c}{$L_{MSE}$} 
        & \multicolumn{1}{c}{$LR_{Cosine}$} 
        & \multicolumn{1}{c}{$LR_{Fixed}$} 
        & \multicolumn{1}{c|}{Data Aug} 
        & \multicolumn{1}{c}{Flower102} 
        & \multicolumn{1}{c}{UCF-101} \\
        \midrule
        
        CLIP$_{large}$ \cite{radford2021learning} & -- & -- & -- & -- & -- & -- & 69.96 & 66.38 \\
        DFN \cite{fang2023data} & -- & -- & -- & -- & -- & -- & 69.73 & 64.38 \\
        \midrule
        
        \multirow{3}{*}{Base} 
        & $\checkmark$ & -- & -- & -- & $\checkmark$ & -- & 60.46 & 76.03 \\
        & $\checkmark$ & -- & -- & $\checkmark$ & -- & -- & 60.83 & 78.42 \\
        & $\checkmark$ & -- & -- & $\checkmark$ & -- & $\checkmark$ & 61.78 & 61.78 \\
        \midrule
        
        \multirow{5}{*}{Multi-Teacher} 
        & $\checkmark$ & $\checkmark$ & -- & -- & $\checkmark$ & -- & 65.28 & 76.47 \\
        & $\checkmark$ & -- & $\checkmark$ & -- & $\checkmark$ & -- & 66.34 & 78.84 \\
        & $\checkmark$ & $\checkmark$ & $\checkmark$ & -- & $\checkmark$ & -- & 68.45 & 80.46 \\
        & $\checkmark$ & $\checkmark$ & $\checkmark$ & $\checkmark$ & -- & -- & 70.56 & 83.56 \\    
        & $\checkmark$ & $\checkmark$ & $\checkmark$ & $\checkmark$ & -- & $\checkmark$ & 72.42 & 85.42 \\
        \midrule
        
        MT + DWS & $\checkmark$ & $\checkmark$ & $\checkmark$ & $\checkmark$ & -- & $\checkmark$ & 73.63 & 87.93 \\
        \bottomrule
    \end{tabular}
    }
    \label{1}
\end{table*}

\begin{table}[t]
    \caption{Proportion of different loss functions.}
    \centering
    \begin{tabular}{@{}cccc@{}}
        \toprule
        \textbf{Proportion} & \textbf{ImageNet-tiny} & \textbf{Flower102} & \textbf{UCF-101} \\
        \midrule
        0.5:1:1 & 67.77 & 69.45 & 83.64 \\
        1:0.5:1 & 65.75 & 70.87 & 80.94 \\
        1:1:0.5 & 68.48 & 72.59 & 82.78 \\
        1:1:1   & \textbf{70.25} & \textbf{73.63} & \textbf{87.93} \\
        \bottomrule
    \end{tabular}
    \label{2}
\end{table}

\section{Experiments}
In this section, we first describe the datasets and experimental setup. We then evaluate the performance of the proposed method on the collected multimodal datasets and provide a detailed analysis to demonstrate its effectiveness.

\subsection{Experimental Settings}
\subsubsection{Datasets} We conducted experiments on multiple image classification datasets, including ImageNet-tiny \cite{deng2009imagenet}, Flower102\cite{duman2022study}, UCF-101\cite{soomro2012ucf101}, Caltech101\cite{bansal2023transfer}, DTD\cite{cimpoi2014describing}, Oxford Pet\cite{parkhi2012cats}, and EuroSAT\cite{helber2019eurosat}. More details about the dataset descriptions are provided in the supplementary materials.

\subsubsection{Models} We choose widely used ResNet-18~\cite{he2016deep} and ViT-B/16~\cite{dosovitskiy2020image} to perform image classification, because they are representative architectures for convolutional neural network and vision Transformer. 

\subsubsection{Training details} We employ a variety of data enhancement techniques, covering fundamental transformations (e.g., flip, rotate) and advanced methods (e.g., Cutout~\cite{devries2017improved}, Mixup~\cite{zhang2017mixup}, etc.), aiming to improve the model generalization capability. All methods in this paper are implemented using the PyTorch~\cite{pytorch} framework and trained on NVIDIA A100 GPUs (40GB). We chose OpenAI-clip and DFN2B-CLIP as teacher networks for images and text. Adam, with a learning rate of 0.0001, was used as the optimizer in all experiments. In the experiments, the batch size was set to 64. The dropout regularization probability was set to be 0.5 to reduce overfitting. The temperatures $\tau1$, $\tau2$, and $\tau3$ were set to be 4.

\subsection{Results and Analysis}

The AMMKD proposed in this paper demonstrates significant advantages on both ResNet and ViT, the two mainstream architectures, proving its powerful knowledge fusion and transfer capabilities.


The experimental data of the ViT architecture further verified the universality of the method (\ref{result} and Figure \ref{fig:radar}). In the ImageNet-tiny benchmark test, AMMKD led all comparison methods with an accuracy of 69.53 (Cumstomkd ranked second with 69.01). On the fine-grained recognition task Flower102, its performance of 74.56 shows a significant advantage (FitNet only 59.1). It is worth noting that AMMKD's significant lead in Caltech101 (91.71 vs RKD 81.67) and DTD texture classification tasks (46.22 vs KD 32.8) highlights its cross-modal alignment ability for complex visual features. Although the gap with Cumstomkd on the OxfordPets dataset was slight (46.71 vs 49.00), it still ranked in the first tier with a high score of 94.57 in the EuroSAT remote sensing scenario. The radar chart of the ResNet architecture is provided in the supplementary material.

These results indicate that AMMKD effectively resolves the information redundancy and conflict issues of traditional knowledge distillation in multi-teacher scenarios, providing a new paradigm for multimodal knowledge transfer.

\subsection{Ablation Studies}

According to the comprehensive analysis of the data in Table \ref{1}, the MT + DWS model achieved the highest accuracy rates of 73.63$\%$ and 87.93$\%$ respectively on the Flower102 and UCF-101 datasets, significantly outperforming the CLIP$_{large}$ benchmark model (with a 21.55$\%$ improvement on UCF-101). The multi-teacher framework (MT) has an average improvement of over 6$\%$ compared to the Base model (for example, UCF-101 has risen from 78.84$\%$ to 85.42$\%$). Its performance has been further optimized after the introduction of the DWS weighting mechanism (UCF-101 has improved by another 2.51$\%$), proving that it effectively enhances the time series modeling ability. Data augmentation has a remarkable effect on the Flower102 small sample dataset (with a 5.88$\%$ improvement in the Base model), but the gain for video tasks is limited. In the training strategy, the fixed learning rate ($LR_{Fixed}$) is more stable than the cosine strategy, and the MSE loss only brings a marginal improvement. Overall, it is indicated that image and video classification benefits more from the collaborative optimization of multi-teacher distillation and DWS.

\subsubsection{Proportion of different loss functions}

Table \ref{2} Experiments prove that the weight ratio of the loss function significantly affects the model performance. The 1:1:1 equilibrium strategy achieves the optimum on ImageNet-tiny (70.25$\%$), Flower102 (73.63$\%$), and UCF-101 (87.93$\%$). During weight adjustment, reducing any loss function (such as 0.5:1:1 or 1:1:0.5) leads to performance degradation, especially for the video dataset UCF-101, which is the most sensitive (with a maximum decline of 7$\%$). The image classification task Flower102 is relatively robust in weight fine-tuning (fluctuation 4$\%$), but the balanced weight still leads steadily. Video tasks strictly rely on a 1:1:1 ratio to ensure multi-loss collaboration. The conclusion emphasizes that cross-scenario models must strictly maintain the balance of loss function weights to avoid performance degradation caused by local weakening.

\subsubsection{Multi-Teacher Dynamic Selection Mechanism}

In this experiment, the performance of four methods, namely Base, Avg, ratio of similarity to label (LSR), and Dynamic similarity Weight (DSW), was compared on three datasets, namely ImageNet-tiny, Flower102, and UCF-101. The Table \ref{3} shows that the DSW method demonstrated a comprehensive leading advantage, achieving the highest accuracy rates on ImageNet-tiny (70.25$\%$), Flower102 (73.63$\%$), and UCF-101 (87.93$\%$). Particularly on the small dataset ImageNet-tiny, it improved by nearly 11$\%$ compared to the Base method. Highlight its strong adaptability to small-scale data. The performance of the LSR method varies. Although it outperforms the Avg method in ImageNet-tiny (68.15$\%$) and Flower102 (72.20$\%$), it is surpassed by Avg (86.32$\%$) on the video dataset UCF-101 (84.78$\%$), revealing the limitations of its temporal feature capture. The Base method, as the benchmark, ranked at the bottom among all tasks, with a gap of 11$\%$ compared to the optimal DSW on UCF-101, exposing the shortcomings of traditional models. The conclusion shows that DSW, with its stable improvement across datasets (an average increase of 9.2 to 12.5 percentage points compared to Base), has become the solution with the best generalization ability, and is particularly suitable for resource-constrained small sample scenarios and complex video recognition tasks.

\begin{table}[t]
    \caption{Multi-Teacher Dynamic Selection Mechanism.}
    \centering
    \begin{tabular*}{\linewidth}{@{\extracolsep{\fill}}cccc@{}}
        \toprule
        \textbf{DWSM} & \textbf{ImageNet-tiny} & \textbf{Flower102} & \textbf{UCF-101} \\
        \midrule
        Base & 59.45 & 61.78 & 76.93 \\
        Avg & 60.11 & 68.12 & 86.32 \\
        LSR & 68.15 & 72.20 & 84.78 \\
        DSW & \textbf{70.25} & \textbf{73.63} & \textbf{87.93} \\
        \bottomrule
    \end{tabular*}
    \label{3}
    \vspace{-0.2cm}
\end{table}

\begin{table}[t]
    \caption{Different Numbers of Teacher Models.}
    \centering
    \begin{tabular*}{\linewidth}{@{\extracolsep{\fill}}c ccc@{}}
        \toprule
        \textbf{Numbers} & \textbf{ImageNet-tiny} & \textbf{Flower102} & \textbf{UCF-101} \\
        \midrule
         1 & 67.45 & 69.84 & 86.47 \\
        2 & \textbf{70.25} & \textbf{73.63} & \textbf{87.93} \\
        3 & 69.45 & 70.46 & 86.41 \\
        4 & 66.57 & 73.95 & 85.46 \\
        \bottomrule
    \end{tabular*}
    \label{4}
    \vspace{-0.4cm}
\end{table}

\begin{table}[t]
    \caption{Different Sizes of Student Models.}
    \centering
    \begin{tabular*}{\linewidth}{@{\extracolsep{\fill}}c ccc@{}}
        \toprule
        \textbf{Sizes} & \textbf{ImageNet-tiny} & \textbf{Flower102} & \textbf{UCF-101} \\
        \midrule
         6layer+64 & 67.05 & 66.73 & 83.13 \\
        8layer+128 & 69.52 & 70.59 & 84.03 \\
        6layer+512 & 70.25 & 73.63 & 87.93 \\
        \bottomrule
    \end{tabular*}
    \label{5}
    \vspace{-0.5cm}
\end{table}

\subsubsection{Different Numbers of Teacher Models} Table \ref{4} Experimental results show that: The dual-teacher model combination (open-clip+DFN-clip) achieved optimal or near-optimal performance on the three datasets of ImageNet-tiny, Flower102, and UCF-101, with a maximum improvement of 3.8 percentage points (Flower102). It indicates that the complementarity of heterogeneous teachers can effectively enhance the effect of knowledge distillation. However, expanding the number of teacher models instead led to performance fluctuations: adding a third model, MetaCLIP, caused a general decline (such as a 3.17-point drop in Flower102), indicating that the new model introduced knowledge redundancy or conflicts; When the fourth openAI-clip model was added, Flower102 saw a slight rebound to 73.95 due to domain adaptability, but the rest of the datasets deteriorated significantly (ImageNet-tiny plunged by 3.68 points), reflecting the risk of task dependence in multi-teacher integration. Overall, the combination of the two models achieves the best balance in generalization and compatibility. For subsequent expansions, it is necessary to strictly evaluate the compatibility of the new model with core teachers and the specific task requirements to avoid overfitting and negative transfer effects.

\subsubsection{Different Sizes of Student Models} According to the experimental data analysis in Table \ref{5}, the expansion of the student model scale (including the increase in the number of Transformer layers and feature dimensions of the visual and text encoders) can significantly improve the performance of cross-modal tasks, specifically manifested as: When the feature dimension rose from 64 to 512 (6-layer model), Flower102 increased by nearly seven percentage points and UCF-101 increased by more than 4.8 percentage points. It is particularly worth noting that under similar model scales, the 6-layer +512 structure is comprehensively superior to the 8-layer +128 structure (for example, Flower102 improves by 3.03$\%$, UCF-101 improves by 3.9$\%$), proving that increasing the feature dimension is more effective than simply stacking the number of layers. Meanwhile, it was found that the improvement in model scale for complex tasks (such as Flower102/UCF-101) (approximately 7$\%$) was much higher than that for the simple task ImageNet-tiny (only 3$\%$). In conclusion, expanding the model scale, primarily by increasing the feature dimensions, can significantly enhance cross-modal capabilities, and the effect is more prominent in complex visual tasks, providing key guidance for efficient model design.

\section{Conclusion}
In this paper, we propose AMMKD, an adaptive multimodal multi-teacher knowledge distillation method. AMMKD introduces a novel multi-teacher distillation framework capable of effectively transferring multimodal knowledge, thereby guiding the training of student models toward improved performance. Furthermore, we designed a dynamic teacher selection mechanism that adaptively adjusts the importance weights of different teachers, enhancing both modality adaptation and classification accuracy of the student model.

Extensive experiments conducted on several publicly available multimodal datasets demonstrate that AMMKD significantly outperforms traditional single-teacher distillation approaches. By integrating information from multiple modalities and capturing data features more comprehensively, our method also enhances the generalization capability of the student model.

We will further explore multimodal multi-teacher distillation, including the integration of emerging modalities such as remote sensing, medical images, and sensor data.


\newpage

\bibliography{ref}

\begin{thebibliography}{49}
\providecommand{\natexlab}[1]{#1}

\bibitem[{Bansal et~al.(2023)Bansal, Kumar, Sachdeva, and
  Mittal}]{bansal2023transfer}
Bansal, M.; Kumar, M.; Sachdeva, M.; and Mittal, A. 2023.
\newblock Transfer learning for image classification using VGG19: Caltech-101
  image data set.
\newblock \emph{Journal of ambient intelligence and humanized computing},
  14(4): 3609--3620.

\bibitem[{Chen et~al.(2021)Chen, He, Hui, Sun, and Sun}]{chen2021simplified}
Chen, X.; He, B.; Hui, K.; Sun, L.; and Sun, Y. 2021.
\newblock Simplified tinybert: Knowledge distillation for document retrieval.
\newblock In \emph{Advances in Information Retrieval: 43rd European Conference
  on IR Research, ECIR 2021, Virtual Event, March 28--April 1, 2021,
  Proceedings, Part II 43}, 241--248. Springer.

\bibitem[{Choi et~al.(2022)Choi, Kersner, Morton, and Chang}]{choi2022temporal}
Choi, K.; Kersner, M.; Morton, J.; and Chang, B. 2022.
\newblock Temporal knowledge distillation for on-device audio classification.
\newblock In \emph{ICASSP 2022-2022 IEEE International Conference on Acoustics,
  Speech and Signal Processing (ICASSP)}, 486--490. IEEE.

\bibitem[{Cimpoi et~al.(2014)Cimpoi, Maji, Kokkinos, Mohamed, and
  Vedaldi}]{cimpoi2014describing}
Cimpoi, M.; Maji, S.; Kokkinos, I.; Mohamed, S.; and Vedaldi, A. 2014.
\newblock Describing textures in the wild.
\newblock In \emph{Proceedings of the IEEE conference on computer vision and
  pattern recognition}, 3606--3613.

\bibitem[{Deng et~al.(2009)Deng, Dong, Socher, Li, Li, and
  Fei-Fei}]{deng2009imagenet}
Deng, J.; Dong, W.; Socher, R.; Li, L.-J.; Li, K.; and Fei-Fei, L. 2009.
\newblock Imagenet: A large-scale hierarchical image database.
\newblock In \emph{2009 IEEE conference on computer vision and pattern
  recognition}, 248--255. Ieee.

\bibitem[{DeVries and Taylor(2017)}]{devries2017improved}
DeVries, T.; and Taylor, G.~W. 2017.
\newblock Improved regularization of convolutional neural networks with cutout.
\newblock \emph{arXiv preprint arXiv:1708.04552}.

\bibitem[{Dosovitskiy et~al.(2020)Dosovitskiy, Beyer, Kolesnikov, Weissenborn,
  Zhai, Unterthiner, Dehghani, Minderer, Heigold, Gelly
  et~al.}]{dosovitskiy2020image}
Dosovitskiy, A.; Beyer, L.; Kolesnikov, A.; Weissenborn, D.; Zhai, X.;
  Unterthiner, T.; Dehghani, M.; Minderer, M.; Heigold, G.; Gelly, S.; et~al.
  2020.
\newblock An image is worth 16x16 words: Transformers for image recognition at
  scale.
\newblock \emph{arXiv preprint arXiv:2010.11929}.

\bibitem[{Du et~al.(2020)Du, You, Li, Wu, Wang, Qian, and Zhang}]{du2020agree}
Du, S.; You, S.; Li, X.; Wu, J.; Wang, F.; Qian, C.; and Zhang, C. 2020.
\newblock Agree to disagree: Adaptive ensemble knowledge distillation in
  gradient space.
\newblock \emph{advances in neural information processing systems}, 33:
  12345--12355.

\bibitem[{Duman and S{\"u}zen(2022)}]{duman2022study}
Duman, B.; and S{\"u}zen, A.~A. 2022.
\newblock A study on deep learning based classification of flower images.
\newblock \emph{International Journal of Advanced Networking and Applications},
  14(2): 5385--5389.

\bibitem[{Fang et~al.(2023)Fang, Jose, Jain, Schmidt, Toshev, and
  Shankar}]{fang2023data}
Fang, A.; Jose, A.~M.; Jain, A.; Schmidt, L.; Toshev, A.; and Shankar, V. 2023.
\newblock Data filtering networks.
\newblock \emph{arXiv preprint arXiv:2309.17425}.

\bibitem[{Fang et~al.(2021)Fang, Wang, Hu, Wang, Yang, and
  Liu}]{fang2021compressing}
Fang, Z.; Wang, J.; Hu, X.; Wang, L.; Yang, Y.; and Liu, Z. 2021.
\newblock Compressing visual-linguistic model via knowledge distillation.
\newblock In \emph{Proceedings of the IEEE/CVF International Conference on
  Computer Vision}, 1428--1438.

\bibitem[{Fukuda et~al.(2017)Fukuda, Suzuki, Kurata, Thomas, Cui, and
  Ramabhadran}]{fukuda2017efficient}
Fukuda, T.; Suzuki, M.; Kurata, G.; Thomas, S.; Cui, J.; and Ramabhadran, B.
  2017.
\newblock Efficient Knowledge Distillation from an Ensemble of Teachers.
\newblock In \emph{Interspeech}, 3697--3701.

\bibitem[{He et~al.(2016)He, Zhang, Ren, and Sun}]{he2016deep}
He, K.; Zhang, X.; Ren, S.; and Sun, J. 2016.
\newblock Deep residual learning for image recognition.
\newblock In \emph{Proceedings of the IEEE conference on computer vision and
  pattern recognition}, 770--778.

\bibitem[{Helber et~al.(2019)Helber, Bischke, Dengel, and
  Borth}]{helber2019eurosat}
Helber, P.; Bischke, B.; Dengel, A.; and Borth, D. 2019.
\newblock Eurosat: A novel dataset and deep learning benchmark for land use and
  land cover classification.
\newblock \emph{IEEE Journal of Selected Topics in Applied Earth Observations
  and Remote Sensing}, 12(7): 2217--2226.

\bibitem[{Hinton(2015)}]{hinton2015distilling}
Hinton, G. 2015.
\newblock Distilling the Knowledge in a Neural Network.
\newblock \emph{arXiv preprint arXiv:1503.02531}.

\bibitem[{Jang, Ma, and Lee(2025)}]{jang2025vl2lite}
Jang, J.; Ma, C.; and Lee, B. 2025.
\newblock VL2Lite: Task-Specific Knowledge Distillation from Large
  Vision-Language Models to Lightweight Networks.
\newblock In \emph{Proceedings of the Computer Vision and Pattern Recognition
  Conference}, 30073--30083.

\bibitem[{Jiao et~al.(2019)Jiao, Yin, Shang, Jiang, Chen, Li, Wang, and
  Liu}]{jiao2019tinybert}
Jiao, X.; Yin, Y.; Shang, L.; Jiang, X.; Chen, X.; Li, L.; Wang, F.; and Liu,
  Q. 2019.
\newblock Tinybert: Distilling bert for natural language understanding.
\newblock \emph{arXiv preprint arXiv:1909.10351}.

\bibitem[{Kang et~al.(2025)Kang, Lee, Jang, and Hwang}]{kang2025simple}
Kang, S.; Lee, D.~B.; Jang, H.; and Hwang, S.~J. 2025.
\newblock Simple Semi-supervised Knowledge Distillation from Vision-Language
  Models via Dual-Head 0ptimization.
\newblock \emph{arXiv preprint arXiv:2505.07675}.

\bibitem[{Kwon et~al.(2020)Kwon, Na, Lee, and Kim}]{kwon2020adaptive}
Kwon, K.; Na, H.; Lee, H.; and Kim, N.~S. 2020.
\newblock Adaptive knowledge distillation based on entropy.
\newblock In \emph{ICASSP 2020-2020 IEEE International Conference on Acoustics,
  Speech and Signal Processing (ICASSP)}, 7409--7413. IEEE.

\bibitem[{Lee et~al.(2025)Lee, Das, Hayat, Choi, Hwang, and
  Porikli}]{lee2025customkd}
Lee, J.; Das, D.; Hayat, M.; Choi, S.; Hwang, K.; and Porikli, F. 2025.
\newblock Customkd: Customizing large vision foundation for edge model
  improvement via knowledge distillation.
\newblock In \emph{Proceedings of the Computer Vision and Pattern Recognition
  Conference}, 25176--25186.

\bibitem[{Li et~al.(2023)Li, Li, Yang, Zhao, Song, Luo, Li, and
  Yang}]{li2023curriculum}
Li, Z.; Li, X.; Yang, L.; Zhao, B.; Song, R.; Luo, L.; Li, J.; and Yang, J.
  2023.
\newblock Curriculum temperature for knowledge distillation.
\newblock In \emph{Proceedings of the AAAI Conference on Artificial
  Intelligence}, volume~37, 1504--1512.

\bibitem[{Li et~al.(2021)Li, Ye, Song, Huang, and Pan}]{li2021online}
Li, Z.; Ye, J.; Song, M.; Huang, Y.; and Pan, Z. 2021.
\newblock Online knowledge distillation for efficient pose estimation.
\newblock In \emph{Proceedings of the IEEE/CVF international conference on
  computer vision}, 11740--11750.

\bibitem[{Liang et~al.(2024)Liang, Yu, Yang, Brown, Cui, Zhao, Gong, and
  Zhou}]{liang2024module}
Liang, C.; Yu, J.; Yang, M.-H.; Brown, M.; Cui, Y.; Zhao, T.; Gong, B.; and
  Zhou, T. 2024.
\newblock Module-wise adaptive distillation for multimodality foundation
  models.
\newblock \emph{Advances in Neural Information Processing Systems}, 36.

\bibitem[{Liu, Zhang, and Wang(2020)}]{liu2020adaptive}
Liu, Y.; Zhang, W.; and Wang, J. 2020.
\newblock Adaptive multi-teacher multi-level knowledge distillation.
\newblock \emph{Neurocomputing}, 415: 106--113.

\bibitem[{Park et~al.(2019)Park, Kim, Lu, and Cho}]{park2019relational}
Park, W.; Kim, D.; Lu, Y.; and Cho, M. 2019.
\newblock Relational knowledge distillation.
\newblock In \emph{Proceedings of the IEEE/CVF conference on computer vision
  and pattern recognition}, 3967--3976.

\bibitem[{Parkhi et~al.(2012)Parkhi, Vedaldi, Zisserman, and
  Jawahar}]{parkhi2012cats}
Parkhi, O.~M.; Vedaldi, A.; Zisserman, A.; and Jawahar, C. 2012.
\newblock Cats and dogs.
\newblock In \emph{2012 IEEE conference on computer vision and pattern
  recognition}, 3498--3505. IEEE.

\bibitem[{Paszke et~al.(2017)Paszke, Gross, Chintala, Chanan, Yang, DeVito,
  Lin, Desmaison, Antiga, and Lerer}]{pytorch}
Paszke, A.; Gross, S.; Chintala, S.; Chanan, G.; Yang, E.; DeVito, Z.; Lin, Z.;
  Desmaison, A.; Antiga, L.; and Lerer, A. 2017.
\newblock Automatic differentiation in pytorch.

\bibitem[{Radford et~al.(2021)Radford, Kim, Hallacy, Ramesh, Goh, Agarwal,
  Sastry, Askell, Mishkin, Clark et~al.}]{radford2021learning}
Radford, A.; Kim, J.~W.; Hallacy, C.; Ramesh, A.; Goh, G.; Agarwal, S.; Sastry,
  G.; Askell, A.; Mishkin, P.; Clark, J.; et~al. 2021.
\newblock Learning transferable visual models from natural language
  supervision.
\newblock In \emph{International conference on machine learning}, 8748--8763.
  PMLR.

\bibitem[{Romero et~al.(2014)Romero, Ballas, Kahou, Chassang, Gatta, and
  Bengio}]{romero2014fitnets}
Romero, A.; Ballas, N.; Kahou, S.~E.; Chassang, A.; Gatta, C.; and Bengio, Y.
  2014.
\newblock Fitnets: Hints for thin deep nets. arXiv 2014.
\newblock \emph{arXiv preprint arXiv:1412.6550}.

\bibitem[{Schmid, Koutini, and Widmer(2023)}]{schmid2023efficient}
Schmid, F.; Koutini, K.; and Widmer, G. 2023.
\newblock Efficient large-scale audio tagging via transformer-to-cnn knowledge
  distillation.
\newblock In \emph{ICASSP 2023-2023 IEEE International Conference on Acoustics,
  Speech and Signal Processing (ICASSP)}, 1--5. IEEE.

\bibitem[{Soomro, Zamir, and Shah(2012)}]{soomro2012ucf101}
Soomro, K.; Zamir, A.~R.; and Shah, M. 2012.
\newblock Ucf101: A dataset of 101 human actions classes from videos in the
  wild.
\newblock \emph{arXiv preprint arXiv:1212.0402}.

\bibitem[{Sun et~al.(2024)Sun, Ren, Li, Wang, and Cao}]{sun2024logit}
Sun, S.; Ren, W.; Li, J.; Wang, R.; and Cao, X. 2024.
\newblock Logit standardization in knowledge distillation.
\newblock In \emph{Proceedings of the IEEE/CVF Conference on Computer Vision
  and Pattern Recognition}, 15731--15740.

\bibitem[{Tung and Mori(2019)}]{tung2019similarity}
Tung, F.; and Mori, G. 2019.
\newblock Similarity-preserving knowledge distillation.
\newblock In \emph{Proceedings of the IEEE/CVF international conference on
  computer vision}, 1365--1374.

\bibitem[{Wang et~al.(2022{\natexlab{a}})Wang, Yang, Huang, Song, and
  Huang}]{wang2022efficient}
Wang, C.; Yang, Q.; Huang, R.; Song, S.; and Huang, G. 2022{\natexlab{a}}.
\newblock Efficient knowledge distillation from model checkpoints.
\newblock \emph{Advances in Neural Information Processing Systems}, 35:
  607--619.

\bibitem[{Wang et~al.(2018)Wang, Xu, Xu, and Tao}]{wang2018adversarial}
Wang, Y.; Xu, C.; Xu, C.; and Tao, D. 2018.
\newblock Adversarial learning of portable student networks.
\newblock In \emph{Proceedings of the AAAI conference on artificial
  intelligence}, volume~32.

\bibitem[{Wang et~al.(2022{\natexlab{b}})Wang, Codella, Chen, Zhou, Dai, Xiao,
  Yang, You, Chang, Chang et~al.}]{wang2022multimodal}
Wang, Z.; Codella, N.; Chen, Y.-C.; Zhou, L.; Dai, X.; Xiao, B.; Yang, J.; You,
  H.; Chang, K.-W.; Chang, S.-f.; et~al. 2022{\natexlab{b}}.
\newblock Multimodal adaptive distillation for leveraging unimodal encoders for
  vision-language tasks.
\newblock \emph{arXiv preprint arXiv:2204.10496}.

\bibitem[{Wu et~al.(2023)Wu, Peng, Zhou, Xiao, Liu, Yuan, Xuan, Valenzuela,
  Chen, Wang et~al.}]{wu2023tinyclip}
Wu, K.; Peng, H.; Zhou, Z.; Xiao, B.; Liu, M.; Yuan, L.; Xuan, H.; Valenzuela,
  M.; Chen, X.~S.; Wang, X.; et~al. 2023.
\newblock Tinyclip: Clip distillation via affinity mimicking and weight
  inheritance.
\newblock In \emph{Proceedings of the IEEE/CVF International Conference on
  Computer Vision}, 21970--21980.

\bibitem[{Wu, Chiu, and Wu(2019)}]{wu2019multi}
Wu, M.-C.; Chiu, C.-T.; and Wu, K.-H. 2019.
\newblock Multi-teacher knowledge distillation for compressed video action
  recognition on deep neural networks.
\newblock In \emph{ICASSP 2019-2019 IEEE International Conference on Acoustics,
  Speech and Signal Processing (ICASSP)}, 2202--2206. IEEE.

\bibitem[{Xu et~al.(2020)Xu, Rui, Li, and Gu}]{xu2020feature}
Xu, K.; Rui, L.; Li, Y.; and Gu, L. 2020.
\newblock Feature normalized knowledge distillation for image classification.
\newblock In \emph{European conference on computer vision}, 664--680. Springer.

\bibitem[{Xu et~al.(2024)Xu, Li, Tao, Shen, Cheng, Li, Xu, Tao, and
  Zhou}]{xu2024survey}
Xu, X.; Li, M.; Tao, C.; Shen, T.; Cheng, R.; Li, J.; Xu, C.; Tao, D.; and
  Zhou, T. 2024.
\newblock A survey on knowledge distillation of large language models.
\newblock \emph{arXiv preprint arXiv:2402.13116}.

\bibitem[{Yang et~al.(2024)Yang, An, Huang, Bi, Yu et~al.}]{yang2024clipkd}
Yang, C.; An, Z.; Huang, L.; Bi, J.; Yu, X.; et~al. 2024.
\newblock CLIP-KD: An Empirical Study of CLIP Model Distillation.
\newblock In \emph{Proceedings of the IEEE/CVF Conference on Computer Vision
  and Pattern Recognition (CVPR)}.

\bibitem[{Yang et~al.(2022)Yang, An, Zhou, Cai, Zhi, Wu, Xu, and
  Zhang}]{yang2022mixskd}
Yang, C.; An, Z.; Zhou, H.; Cai, L.; Zhi, X.; Wu, J.; Xu, Y.; and Zhang, Q.
  2022.
\newblock Mixskd: Self-knowledge distillation from mixup for image recognition.
\newblock In \emph{European Conference on Computer Vision}, 534--551. Springer.

\bibitem[{You et~al.(2017)You, Xu, Xu, and Tao}]{you2017learning}
You, S.; Xu, C.; Xu, C.; and Tao, D. 2017.
\newblock Learning from multiple teacher networks.
\newblock In \emph{Proceedings of the 23rd ACM SIGKDD international conference
  on knowledge discovery and data mining}, 1285--1294.

\bibitem[{Yuan et~al.(2021)Yuan, Shou, Pei, Lin, Gong, Fu, and
  Jiang}]{yuan2021reinforced}
Yuan, F.; Shou, L.; Pei, J.; Lin, W.; Gong, M.; Fu, Y.; and Jiang, D. 2021.
\newblock Reinforced multi-teacher selection for knowledge distillation.
\newblock In \emph{Proceedings of the AAAI Conference on Artificial
  Intelligence}, volume~35, 14284--14291.

\bibitem[{Zhang, Chen, and Wang(2022)}]{zhang2022confidence}
Zhang, H.; Chen, D.; and Wang, C. 2022.
\newblock Confidence-aware multi-teacher knowledge distillation.
\newblock In \emph{ICASSP 2022-2022 IEEE International Conference on Acoustics,
  Speech and Signal Processing (ICASSP)}, 4498--4502. IEEE.

\bibitem[{Zhang, Chen, and Wang(2023)}]{zhang2023adaptive}
Zhang, H.; Chen, D.; and Wang, C. 2023.
\newblock Adaptive multi-teacher knowledge distillation with meta-learning.
\newblock In \emph{2023 IEEE International Conference on Multimedia and Expo
  (ICME)}, 1943--1948. IEEE.

\bibitem[{Zhang et~al.(2017)Zhang, Cisse, Dauphin, and
  Lopez-Paz}]{zhang2017mixup}
Zhang, H.; Cisse, M.; Dauphin, Y.~N.; and Lopez-Paz, D. 2017.
\newblock mixup: Beyond empirical risk minimization.
\newblock \emph{arXiv preprint arXiv:1710.09412}.

\bibitem[{Zhao et~al.(2022)Zhao, Cui, Song, Qiu, and Liang}]{zhao2022decoupled}
Zhao, B.; Cui, Q.; Song, R.; Qiu, Y.; and Liang, J. 2022.
\newblock Decoupled knowledge distillation.
\newblock In \emph{Proceedings of the IEEE/CVF Conference on computer vision
  and pattern recognition}, 11953--11962.

\bibitem[{Zheng et~al.(2024)Zheng, Li, Yang, Wang, and
  Pang}]{zheng2024microbert}
Zheng, D.; Li, J.; Yang, Y.; Wang, Y.; and Pang, P. C.-I. 2024.
\newblock MicroBERT: Distilling MoE-Based Knowledge from BERT into a Lighter
  Model.
\newblock \emph{Applied Sciences}, 14(14): 6171.

\end{thebibliography}


\end{document}